\documentclass{article}

\usepackage[preprint,nonatbib]{neurips_2020}

\usepackage[utf8]{inputenc} % allow utf-8 input
\usepackage[T1]{fontenc}    % use 8-bit T1 fonts
\usepackage{hyperref}       % hyperlinks
\usepackage{url}            % simple URL typesetting
\usepackage{booktabs}       % professional-quality tables
\usepackage{amsfonts}       % blackboard math symbols
\usepackage{amsmath}
\usepackage{nicefrac}       % compact symbols for 1/2, etc.
\usepackage{microtype}      % microtypography
\usepackage{graphicx}       % include graphs
\usepackage{color}          % text color
\usepackage{amsthm}         % proof env
\usepackage{amssymb}        % additional symbol

\usepackage{todonotes}
\newcommand{\method}{ExNODE\space}
\newcommand{\methodnospace}{ExNODE}
\newcommand{\R}{\mathbb{R}}

\newtheorem{theorem}{Theorem}
\newtheorem{definition}{Definition}
\newtheorem{proposition}{Proposition}

\title{Exchangeable Neural ODE for Set Modeling}

\author{%
   Yang Li\thanks{equal contribution, order determined alphabetically} \\
   Department of Computer Science\\
   UNC-Chapel Hill\\
   \texttt{yangli95@cs.unc.edu} \\
   \And
   Haidong Yi\footnotemark[1]\\
   Department of Computer Science\\
   UNC-Chapel Hill\\
   \texttt{haidyi@cs.unc.edu} \\
   \And
   Christopher M.~Bender \\
   Department of Computer Science\\
   UNC-Chapel Hill\\
   \texttt{bender@cs.unc.edu}
   \And
   Siyuan Shan \\
   Department of Computer Science \\
   UNC-Chapel Hill \\
   \texttt{siyuanshan@cs.unc.edu} \\
   \And
   Junier B.~Oliva \\
   Department of Computer Science\\
   The University of North Carolina\\
   Chapel Hill, NC 27599 \\
   \texttt{joliva@cs.unc.edu} \\
}

\begin{document}

\maketitle

% \begin{abstract}
% Instead of modeling each instance independently, many machine learning tasks require to model a set of instances jointly.
% Any model that operates on set structured data should not depend on the ordering of the set elements, a.k.a \emph{permutation invariant}. In this work, we propose an exchangeable module based on differential equation, which captures the intradependencies within sets. Our proposed module can be seamlessly applied for both discriminative and generative tasks. We also extend set modeling in the temporal dimension and propose a VAE based model for temporal set modeling. Extensive experiments demonstrate the efficacy of our method over strong baselines.
% \end{abstract}

\begin{abstract}
Reasoning over an instance composed of a set of vectors, like a point cloud, requires that one accounts for intra-set dependent features among elements. However, since such instances are unordered, the elements' features should remain unchanged when the input's order is permuted. This property, permutation equivariance, is a challenging constraint for most neural architectures. While recent work has proposed global pooling and attention-based solutions, these may be limited in the way that intradependencies are captured in practice. In this work we propose a more general formulation to achieve permutation equivariance through ordinary differential equations (ODE). Our proposed module, Exchangeable Neural ODE (ExNODE), can be seamlessly applied for both discriminative and generative tasks. We also extend set modeling in the temporal dimension and propose a VAE based model for temporal set modeling. Extensive experiments demonstrate the efficacy of our method over strong baselines.
\end{abstract}

\section{Introduction}

The vast majority of machine learning models operate on an independent and identically distributed (i.i.d.) vector, $x \in \R^d$.
In some cases, however, the inputs may contain a set of instances, $\mathbf{x}=\{x_i\}_{i=1}^n$, which jointly determine the target.
We note that instances within a set may interact with each other. For instance, the points inside a point cloud jointly determine the global structure.
In this work, we build both discriminative and generative models on sets, which explore the intradependencies within a set to capture both global and local structures.

A set is a collection of data that does not possess any inherent ordering of its elements.
In statistics, a set is described as an exchangeable sequence of random variables whose joint probability distribution does not change under any permutation $\pi$, i.e.,
\begin{equation}
    p(x_1,\ldots,x_n) = p(x_{\pi_1},\ldots,x_{\pi_n}).
\end{equation}

Discriminative models that operate on a set must predict a target $y$ that is invariant to all permutations.
Applications for such models include population statistics estimation, point cloud classification, etc.
A naive approach where training data are augmented with random permutations and treated as sequences has been empirically proven insufficient \cite{vinyals2015order}.
Previous works \cite{zaheer2017deep,edwards2016towards} developed simple permutation invariant operations by processing each element independently and then aggregating them using a pooling operation (max, mean, etc).
However, such an operation largely ignores the intradependencies between elements within the set.
In this work, we introduce an inductive bias into the model to exploit said intradependencies across elements.
Specifically, we introduce a permutation equivariant module to explicitly model the dependencies among set elements.

Set generative models with tractable, exchangeable likelihoods have recently been investigated (that is, likelihoods which are invariant to permutations) \cite{edwards2016towards,bruno2018,bender2020}.
Simple approaches that estimate likelihood for each instance independently are insufficient since global structures cannot be inferred.
To overcome this shortcoming, we construct a flow-based generative model for tractable likelihood estimation on sets.
%We show that combining a sequence of equivariant flow transformations and an invariant likelihood results in an invariant likelihood. 

The key for both discriminative and generative set models is a powerful equivariant transformation that captures set intradependencies.
In order to compute the likelihood for a flow based generative model, the transformation additionally requires to be invertible.
In this work, we propose an exchangeable, invertible flow transformation, \methodnospace, based on Neural Ordinary Differential Equation (NODE) \cite{chen2018neural}.
Invertibility is guaranteed via the NODE framework since integration backward in time is always possible.
We implement \method by parametrizing a differential equation with a permutation equivariant architecture.
% We further exploit previous works with exchangeable attention mechanism to learn the dependencies among set elements \cite{lee19d}.

In addition to modeling the sets in spatial dimensions, we extend \method to the temporal dimension and propose a temporal set modeling task.
Such a set model has many potential applications, including modeling the evolution of galaxies, pedestrian tracking, etc.
Here, we utilize a VAE-based framework with our proposed set generative model as the decoder.
The temporal evolution is captured by another ODE in the latent space.
After training, our model can interpolate and extrapolate to generate sets at unseen (potentially fractional) time steps.

Our contributions are as follows:
% We propose \methodnospace, an exchangeable, invertible module for set modeling that can be applied for both discriminative and flow-based generative models.
1) We propose \methodnospace, an exchangeable module for set modeling, which explicitly captures the intradependencies among set elements.
2) \method represents a type of invertible flow transformation on which the invariant set likelihood can be achieved.
3) We propose a temporal set modeling task and a VAE-based model for time variant set modeling. The temporal VAE utilizes differential equations to transit hidden states in time. To the best of our knowledge, our model is the first one designed for temporal sets.
4) We achieve state-of-the-art performance for both point cloud classification and likelihood estimation.

\section{Background and Related Works}
\label{bkd}

\subsection{Set Modeling}
A set is a collection that does not impose ordering among its elements.
Models over sets \emph{must} preserve this property.
We list the commonly used terminology for set modeling below.
We denote a set as $ \mathbf{x} = \{x_i\}_{i=1}^n \in \mathcal{X}^n$, where $n$ is the cardinality of the set and the calligraphic letter $\mathcal{X}$ represents the domain of each element $x_i$.

\begin{definition}
(Permutation Equivariant) Let $f: \mathcal{X}^n \rightarrow \mathcal{Y}^n$ be a function, then $f$ is permutation equivariant iff for any permutation $\pi(\cdot)$, $f(\pi(\mathbf{x})) = \pi(f(\mathbf{x}))$.
\end{definition}

\begin{definition}
(Permutation Invariant) Let $f: \mathcal{X}^n \rightarrow \mathcal{Y}$ be a function, then $f$ is permutation invariant iff for any permutation $\pi(\cdot)$, $f(\pi(\mathbf{x})) = f(\mathbf{x})$.
\end{definition}

A naive method to encourage permutation invariance is to augment the training data with randomly permuted sets and treat them as sequences.
One could then train a neural network mapping the permuted inputs to the same output.
Due to the universal approximation ability of neural network, the final model could be invariant to permutations given an infinite amount of training data and model capacity.
However, this simple approach does not guarantee invariance for real-world finite datasets.
As pointed out in \cite{vinyals2015order}, the order cannot be discarded for a sequence model.

DeepSet \cite{zaheer2017deep} proves that any permutation invariant function for a set with finite number of elements can be decomposed as 
$\rho(\sum_{x \in \mathcal{X}^n} \phi(x))$, where the summation is over the set elements.
Based on this decomposition, they propose using two neural networks for both $\rho$ and $\phi$ to learn flexible permutation invariant functions.
They also propose an equivariant model, where independent processing combined with a pooling operation is used to capture the intradependencies.
The Set Transformer \cite{lee19d} proposes to use an attention mechanism over set elements to model the intradependencies between each pair of elements.
Since the attention is a weighted sum over all set elements, this operation is naturally equivariant.
They also propose an attention based pooling operation to achieve invariant representations.

Set likelihood estimation requires the likelihood to be invariant to permutations, i.e.
\begin{definition}
(Exchangeable Likelihood) Given any permutation $\pi$ and finite random variables $x_i, i=1, \ldots, n$, then the likelihood of $\{x_i\}_{i=1}^n$ is exchangeable iff 
$$p(x_1, \ldots, x_n) = p(x_{\pi_1}, \ldots, x_{\pi_n})$$
\end{definition}
A naive approach might be to estimate the likelihood independently for each element.
Neural Statistician \cite{edwards2016towards} utilizes a VAE-based model inspired by the de Finetti's theorem, where conditionally independent likelihoods are estimated for each element by conditioning on a permutation invariant latent code.
PointFlow \cite{yang2019pointflow} extends the Neural Statistician by using normalizing flow for both the encoder and decoder.
Both flow models operate independently on each element.
BRUNO \cite{bruno2018} employs an independent flow transformation for each element and an exchangeable student-t process for the invariant likelihood.
FlowScan \cite{bender2020} transforms the set likelihood problem to the familiar sequence likelihood problem via a scan sorting operation.
In this work, we extend a flow based generative model for exchangeable sets with a tractable invariant likelihood.

\subsection{Neural ODE} 
Connection between neural networks and differential equations has been studied in \cite{weinan2017proposal,lu2017beyond}, where classic neural network architectures are interpreted as discretizations of differential equations.  Built upon those works, \cite{chen2018neural} proposed the \emph{Neural ODE}, which employs the adjoint method to optimize the model in a memory efficient way.
Based on the connection between ResNet \cite{he2016deep} and Euler discretized ODE solver, they propose to use other ODE solvers to implicitly build more advanced architectures.
A basic formulation of Neural ODE is shown as:
\begin{equation} \label{eq:basic_node}
    \frac{dh(t)}{dt} = f_\theta(h(t), t), \quad h(t_0) = x,
\end{equation}
where $f_\theta$ is parametrized as a neural network. Neural ODE blocks process input $h(t_0)$ using a black-box ODE solver so that
\begin{equation} \label{eq:int_node}
    h(t_1) = h(t_0) + \int_{t_0}^{t_1} f_\theta(h(t),t)dt.
\end{equation}
Neural ODE represents a type of continuous depth neural network.
Comparing with discrete depth neural networks, Neural ODE has several advantages:
  1) In theory, there could be infinite number of layers that share the same set of parameters. Hence, Neural ODE is more parameter efficient.
  2) The gradients w.r.t.~$\theta$ can be computed using adjoint method \cite{chen2018neural} that only requires $O(1)$ memory usage, since the intermediate variables do not need to be stored during forward pass; they can be recovered during back propagation.
  3) Neural ODE is naturally invertible if $f_\theta$ satisfies certain conditions, such as Lipschitz continuity.

\subsection{Continuous Normalizing Flow (CNF)}
Normalizing flows (NFs) \cite{dinh2014nice,dinh2016density,kingma2018glow} are a family of methods for modeling complex distributions in which both sampling and density evaluation can be efficient and exact.
NFs use the change of variable theorem to calculate the likelihood of training data:
\begin{equation}\label{eq:chg-dis}
\log p_{\mathcal{X}}(x)=\log p_{\mathcal{Z}}(z)+\log \left|\operatorname{det}\frac{\partial q(x)}{\partial x}\right|,
\end{equation}
where $p_\mathcal{X}(x)$ is the likelihood in input space, $p_\mathcal{Z}(z)$ is the likelihood evaluated on a base distribution, and $z=q(x)$ is an invertible transformation which transforms inputs to latent space.
The base distribution is typically chosen as a simple distribution such as isotropic Gaussian.
To allow efficient likelihood evaluation, NFs typically employ transformations $q(\cdot)$ with a triangular Jacobian so that the determinants can be computed cheaply, although it reduces the flexibility and capacity of NFs. 

\cite{chen2018neural} and \cite{grathwohl2018scalable} propose continuous normalizing flows (CNFs) and extend the change of variable theorem to continuous-time case: 
\begin{equation} \label{eq:chg_diffeq}
\frac{d \log p(z(t))}{d t}=-\operatorname{Tr}\left(\frac{\partial f}{\partial z(t)}\right),
\end{equation}
where $\frac{dz}{dt} = f(z(t), t)$ is a differential equation describing the dynamics of $z(t)$ as in Eq.~\eqref{eq:basic_node}.
Unlike in Eq.~\eqref{eq:chg-dis} where variables are transformed explicitly by $q$, CNF implicitly transforms the variables by integration, i.e.,
\begin{equation}\label{eq:implicit_trans}
    q(x) = z(t_1) = z(t_0) + \int_{t_0}^{t_1} f(z(t), t)dt, \quad x = z(t_0)
\end{equation}
Eq.~\eqref{eq:chg_diffeq} requires only the trace of Jacobian matrix rather than the more expensive determinants in Eq.~\eqref{eq:chg-dis}, which reduces the computation complexity dramatically.
As a result, CNFs can afford using more flexible transformations implicitly implemented by integrating $f$.
Equation \eqref{eq:chg_diffeq} also indicates that the change of log density is determined by another ODE that can be solved with $z(t)$ itself simultaneously using an ODE solver.

\section{Method}
\label{mtd}

In this section, we introduce the permutation equivariant module, \methodnospace. We discuss how to apply \method for different set modeling tasks.
We consider both discriminative (set classification) and generative (set generation with flow models) tasks.
Finally, we explore temporal set modeling.

\subsection{Exchangeable Neural ODE}
Our permutation equivariant module for exchangeable sets is based on differential equations. Specifically, we can prove the following theorem. The detailed proof is provided in Appendix \ref{sec:proof}.

\begin{theorem}
(Permutation Equivariant ODE) Given an ODE $\dot{\mathbf{z}}(t) = f(\mathbf{z}(t), t), \mathbf{z}(t)\in \mathcal{X}^n$ defined in an interval $[t_1, t_2]$. If function $f(\mathbf{z}(t), t)$ is permutation equivariant w.r.t. $\mathbf{z}(t)$, then the solution of the ODE, i.e., $\mathbf{z}^\star(t), t\in[t_1, t_2]$ is permutation equivariant w.r.t. the initial value $\mathbf{z}(t_1)$. We call the ODE with permutation equivariant properties ExODE.
\end{theorem}

Following Neural ODE \cite{chen2018neural}, we parametrize $\dot{\mathbf{z}}(t)$ with a neural network.
To ensure the integrated function $\mathbf{z}^*(t)$ is permutation equivariant, we build $\dot{\mathbf{z}}(t)$ in a permutation equivariant form.
Specifically, $f$ is implemented as a permutation equivariant neural network, such as the deepset equivariant layer or the attention based set transformer layer.

An additional benefit of our \method is its invetibility. Since we can always integrate from $t_2$ to $t_1$, it does not require any special design as in typical flow models to guarantee invertibility.
Therefore, our \method can be easily plugged into flow models as a transformation. According to Eq.~\eqref{eq:chg_diffeq}, the likelihood can be similarly evaluated. 

\subsection{Set Classification}

\begin{figure}[ht]
  \centering
  \includegraphics[width=0.98\textwidth]{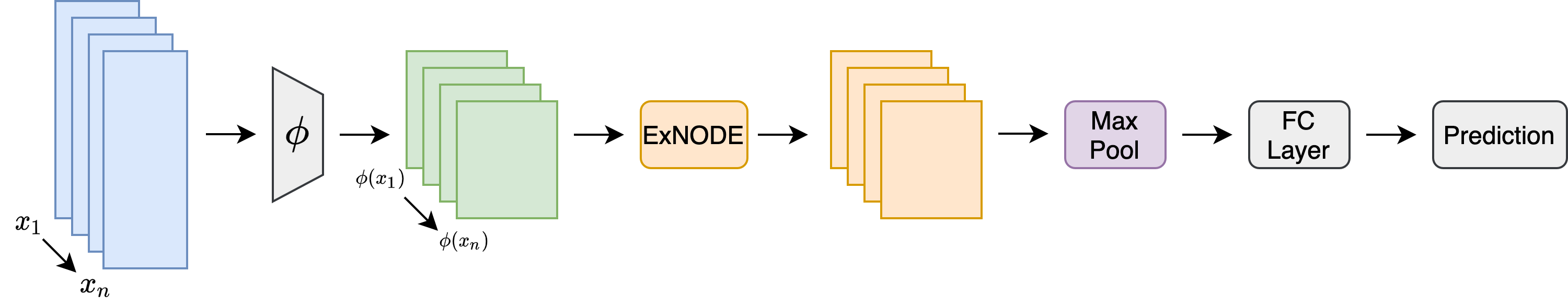}
  \caption{Illustration of the architecture of our set classification model. The function $\phi(\cdot)$ refers to independent operations that expand the dimension. The \emph{ExNODE} may contain multiple ODE blocks. The max pooling is applied across set elements.}
\label{fig:set_classification}
\end{figure}

For the set classification task, a model must guarantee that the order of set elements does not affect the prediction results.
Hence, given a set $\mathbf{x} = \{x_1, \ldots, x_n\} \in \mathcal{X}^n$, our purpose is to learn a permutation invariant function $f(\cdot)$ that maps $\mathbf{x}$ to its corresponding label $y$.

Notice that multiple equivariant layers stacked together are overall equivariant, we employ a permutation invariant architecture by stacking multiple equivariant layers and a pooling aggregating operation.
Figure \ref{fig:set_classification} illustrates the architecture of our set classification model.
First, we use a linear mapping $\phi$ to expand the feature dimensions independently for each set element.
Then, permutation equivariant ODEs serve as a dimension-preserving nonlinear mapping to capture the dependencies among set elements and learn the feature representations for $\mathbf{x}$.
When feature representations are available, we use a max pooling to aggregate the information across $x_i$.
After max pooling, we get a permutation invariant vector representation that summarizes the set $\mathbf{x}$.
We denote the embedding vector as $v$,
\begin{equation}
    v = \mathrm{MaxPool}(\mathrm{ExNODE~Solve}(\phi(\mathbf{x}))).
\end{equation}
Finally, we use fully connected (FC) layers and a softmax layer to predict labels $y$.

\subsection{Continuous Normalizing Flow for Sets}
We extend the continuous normalizing flow proposed in \cite{chen2018neural,grathwohl2018scalable} to model exchangeable sets $\mathbf{x} \in \mathcal{X}^n$.
Specifically, we have the following proposition from \cite{bender2020}, repeated here for convenience:
\begin{proposition}
For a flow model with transformation $q(\cdot)$ and base likelihood $p_{\mathcal{Z}}(\cdot)$, the input likelihood $p_{\mathcal{X}}(\mathbf{x}) = p_{\mathcal{Z}}(q(\mathbf{x})) \left|\det \frac{dq}{d\mathbf{x}}\right|$ is exchangeable iff the transformation is permutation equivariant and the base likelihood is invariant. 
\end{proposition}
Similar to Eq.~\ref{eq:implicit_trans}, we parametrize transformation $q$ implicitly as a differential equation, i.e.,
\begin{equation}\label{eq:int}
    \dot{\mathbf{z}}(t) = f_\theta(\mathbf{z}(t), t), \quad \mathbf{z}(t_0) = \mathbf{x},
\end{equation}
where $f_\theta$ is a permutation equivariant neural network w.r.t. $\mathbf{z}(t)$. 
Using the instantaneous change of variables formula, the log likelihood of $\mathbf{z}(t_1)$ and $\mathbf{z}(t_0)$ satisfy the following equation:
\begin{equation}\label{eq:chg}
	\log p(\mathbf{z}(t_1)) = \log p(\mathbf{z}(t_0)) - \int_{t_{1}}^{t_{0}} \mathrm{Tr}\left(\frac{\partial f_\theta}{\partial \mathbf{z}(t)}\right) dt,
\end{equation}
where $\mathbf{z}(t_0)$ and $\mathbf{z}(t_1)$ corresponds to $x$ and $z$ in Eq.~\eqref{eq:chg-dis} respectively.
Since the trace operator $\mathrm{Tr}(\cdot)$ in Eq.~\eqref{eq:chg} preserves permutation invariance, the exchangeability of $\log p(\mathbf{z}(t))$ is maintained along the integral trajectory.

After transformation, we apply a permutation invariant base likelihood to the transformed sets $\mathbf{z}(t)$. 
For simplicity, we use an i.i.d.~base likelihood
\begin{equation}\label{eq:base_likel}
    p_{\mathcal{Z}}(\mathbf{z}(t)) = \prod_{z_i \in \mathbf{z}(t)} p_{\mathcal{Z}}(z_i).
\end{equation}
The generation process consists of the following steps:
  1) Sampling $n$ i.i.d.~instances from the base distribution;
  2) Inverting the transformations by integrating backwards in time.
  Although samples from base distribution are independent, the transformations will induce dependencies and transform them to encode global and local structures.

\paragraph{Training} Like other normalizing flow based models, we train our model by maximizing the log likelihood $\log p_{\mathcal{X}}(\mathbf{x})$ using Eqs.~\eqref{eq:chg} and \eqref{eq:base_likel}.
We choose $p_{\mathcal{Z}}(\cdot)$ as $\mathcal{N}(0, I)$ in all our experiments.
To reduce memory usage, the adjoint method is used to compute the gradient of a black-box ODE solver \cite{chen2018neural}.
As in FFJORD \cite{grathwohl2018scalable}, the trace of Jacobian matrix is estimated using Hutchinson's estimator \cite{hutchinson1990stochastic}.

\vspace{-5pt}
\subsection{Temporal Set Modeling}
In this section, we present a continuous-time VAE model for temporal set modeling.
Assume $X = [\mathbf{x}_{t_0}, \mathbf{x}_{t_1}, \ldots, \mathbf{x}_{t_N}]$ is a time variant set, where each $\mathbf{x}_{t_i} \in \mathcal{X}^n$ is a set.
Let $Z = [z_{t_0}, z_{t_1}, \ldots, z_{t_N}]$ be the corresponding latent variables of $X$.
We assume that the evolution of latent states can be modeled by an ODE.
In other words, given an initial state $z_{t_0}$, other latent states can be inferred following the dynamics $\dot{z}(t)$.
Unlike other methods, such as recurrent neural networks (RNNs), where the evaluations can only be performed at prefixed time points, the ODE based model can obtain both the latent states and observations at any time $t$. 

Given the latent states $z_{t_i}, i=0,1,\ldots,T$, we propose to model the conditional distribution, $p(\mathbf{x}_{t_i}\mid z_{t_i})$ using a conditional set CNF. 
Specifically, the set $\mathbf{x}_{t_i}$ is transformed to a simple base distribution using \method transformations conditioned on the corresponding latent state $z_{t_i}$:
$$ \mathbf{x}_{t_i}(s_1) = \mathbf{x}_{t_i}(s_0) + \int_{s_0}^{s_1} g_{\theta_d}(\mathbf{x}_{t_i}(s), z_{t_i}, s) ds, \quad \mathbf{x}_{t_i}(s_0) = \mathbf{x}_{t_i},$$
where $g_{\theta_d}(\cdot)$ defines the transformation dynamics of the CNF in $[s_0, s_1]$. $g_{\theta_d}(\cdot)$ is permutation equivariant w.r.t. $\mathbf{x}_{t_i}(s)$.
The log likelihood of $\mathbf{x}_{t_i}$ can be formulated similar to Eq. \eqref{eq:chg}.

\begin{figure}
  \centering
  \includegraphics[width=0.8\linewidth]{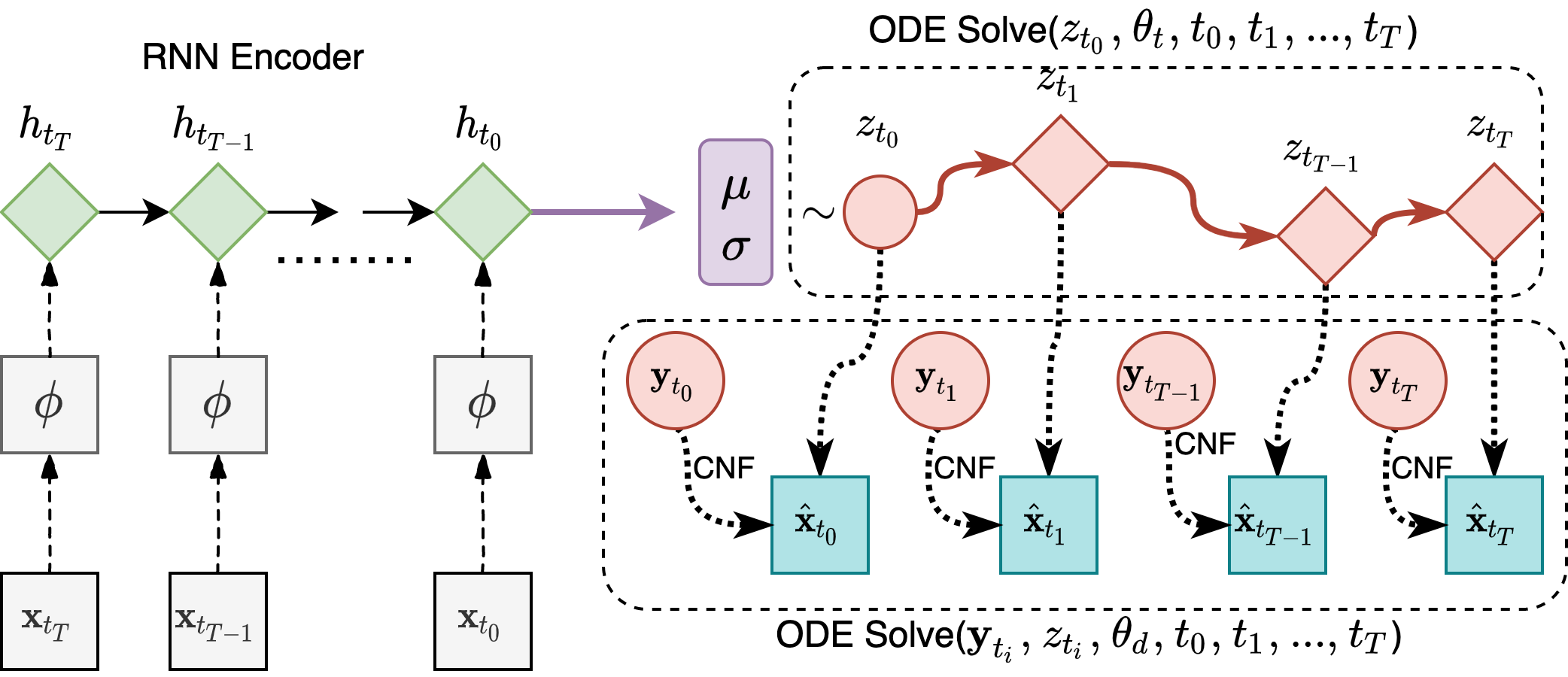}
  \caption{The illustration of encoder and decoder used in temporal set modeling task. The set encoder $\phi(\cdot)$ learns the fix-dimensional permutation invariant representation of a set. The decoder contains two independent ODEs to decode latent states $\mathbf{z}_{t_i}$ and to reconstructed observations $\hat{\mathbf{x}}_{t_i}, i=0,\ldots,T$.} \label{fig:setvae}
\end{figure}

\paragraph{Traning} Since computing the posterior distribution $p(z_{t_i} \mid \mathbf{x}_{t_i})$ is intractable, we cannot directly maximize the marginal log likelihood $\log p_\theta(X)$. Therefore, we resort to the variational inference \cite{VAE2014,rezende2014stochastic} and optimize a lower bound.
Following previous work \cite{chen2018neural,rubanova2019latent} for temporal VAEs, we utilize a recurrent encoder that produces an amortized proposal distribution $\hat{p}_{\psi}(z_{t_0} \mid X)$ conditioned on the entire time series $X$.
The encoder first encodes each set into a permutation invariant representation independently and then uses a recurrent network to accumulate information from each time step.
For our models, the encoder processes the time series backwards in time.
We assume the prior for $z_{t_0}$ comes from an isotropic Gaussian, $p(z_{t_0}) \sim \mathcal{N}(0, I)$.
Latent codes for other time steps are constructed following the dynamics $\dot{z}(t)$.
The final encoder-decoder model is illustrated in Fig.~\ref{fig:setvae}.
We train the encoder and decoder jointly by maximizing the evidence lower bound (ELBO):
\begin{equation}
	\mathrm{ELBO}(\theta, \psi) = \mathbb{E}_{z_{t_0}\sim \hat{p}_\psi(z_{t_0}|X)} \left[ \sum_{i=0}^T \log p_\theta(\mathbf{x}_{t_i}|z_{t_i}) \right ] - \mathrm{KL}(\hat{p}_\psi(z_{t_0}|X) || p(z_{t_0})).
\end{equation}

\paragraph{Sampling}
After the model is trained, we can sample a set at any time $t$ by first inferring the corresponding latent state $z_t$ and then transforming a set of base samples $\mathbf{y}_t$ conditioned on $z_t$:
\begin{gather}
z_{t_{0}} \sim p\left(z_{t_{0}}\right), \quad z_{t} = \mathrm{ODESolve} (z_{t_{0}}, \theta_t, t) \\
\mathbf{y}_t = \{\mathbf{y}_t^j \}_{j=1}^{n}, \quad  \mathbf{y}_t^j \sim \mathcal{N}(0, I), \quad \hat{\mathbf{x}}_t = \mathrm{ODESolve}(\mathbf{y}_t, z_t, \theta_d, t),
\end{gather}
where $\theta_t$ parametrize the dynamics in the latent-states transmission model and $\theta_d$ parametrize the dynamics of the decoder.
Due to the continuous latent space, our model can learn the evolution of sets in time.
We can sample sets at unseen time steps by interpolating or extrapolating the latent states. 
% In addition, comparing with standard RNN based methods that directly learn the likelihood of training data, our encoder $q_\psi$ gives a good uncertainty quantification of the posterior distribution.

\vspace{-5pt}
\section{Experiment}\label{ex}
\vspace{-5pt}

The experiments are divided into three parts.
  First, we evaluate \method on point cloud classification (Sec.~\ref{subsec:pts_cls}). 
  Second, we conduct experiments to validate the efficacy of \method for point cloud generation and likelihood estimation (Sec.~\ref{subsec:set_gen}).
  Finally, we explore the temporal set modeling task (Sec.~\ref{subsec:set_temp}), where interpolated and extrapolated samples are generated to demonstrate the benifits of the continuous-time model.
Our implementation of neural ODE utilizes the official implementation of the NODE \cite{chen2018neural}.
% \footnote{\url{https://github.com/rtqichen/torchdiffeq}}

\vspace{-5pt}
\subsection{Architecture}
\vspace{-5pt}
We consider two different exchangeable base architectures in constructing an \method model: one based on DeepSets \cite{zaheer2017deep} and the other on  Set Transformers \cite{lee19d}.

DeepSets provides both necessary and sufficient conditions for implementing permutation equivariant functions.
In practice, independent element-wise and pooling operations are used to preserve equivariance and capture dependencies, i.e.~$f(\mathbf{x}) = \sigma(\lambda I \mathbf{x} + \gamma ~\text{pool}(\mathbf{x}))$.

Recently, the attention-based \emph{Transformer} has remarkably boosted performance in natural language processing since the transformer can encode pair-wise interactions between elements in sequential data \cite{vaswani2017attention}.
Set Transformer extends the transformer architecture to sets by defining self-attention based operations over set elements.
Using self-attention mechanism comes with several advantages:
  1) pair-wise interactions are explicitly modeled;
  2) stacking multiple blocks can capture higher-order interactions.

\vspace{-5pt}
\subsection{PointCloud Classification}\label{subsec:pts_cls}
\vspace{-5pt}

In this section, we evaluate \method on point cloud classification using  ModelNet40 \cite{wu20153d}, which is composed of surface points from 40 different categories of 3D CAD models.
We train our model with randomly sampled 100 and 1000 points, respectively.
Since \method can easily utilize different exchangeable blocks to learn set representations, we train our model using both DeepSets and Set Transformer blocks.
The test classification accuracy of different models are shown in Table \ref{tab:pts}. We report the mean and standard deviation from 5 runs with different random seeds.
For both small sets (100pts) and large sets (1000pts), \method consistently outperforms the baselines.
\method additionally achieves better performance in terms of parameter efficiency, requiring approximately half the number of parameters compared to Set Transformer and still achieving superior performance.

\begin{table}[t]
    \caption{Test Accuracy for point cloud classification with 100 and 1000 points of ModelNet40 dataset. Mean and standard deviation is reported from 5 runs.}
    \label{tab:pts}
    \centering
    \begin{tabular}{lccc}
    \toprule
        Method                            &      100pts          &   1000pts         &   \# Params  \\ \midrule 
        DeepSets \cite{zaheer2017deep} & 0.82 $\pm$ 0.02      & 0.87   $\pm$ 0.01    &  0.21 M \\
        Set Transformer \cite{lee19d}   & 0.8454 $\pm$ 0.0144  & 0.8915 $\pm$ 0.0144  & 1.15 M \\
        \method (deepset block)   & $\mathbf{0.8597}\pm\mathbf{0.0027}$   & $0.8881\pm0.0016$  & 0.58 M \\
        \method (transformer block)   & $0.8569\pm0.0015$  & $\mathbf{0.8932}\pm\mathbf{0.004}$  & 0.52 M \\
    \bottomrule
    \end{tabular}
\end{table}

\begin{figure}[t]
    \centering
    \includegraphics[width=0.95\linewidth]{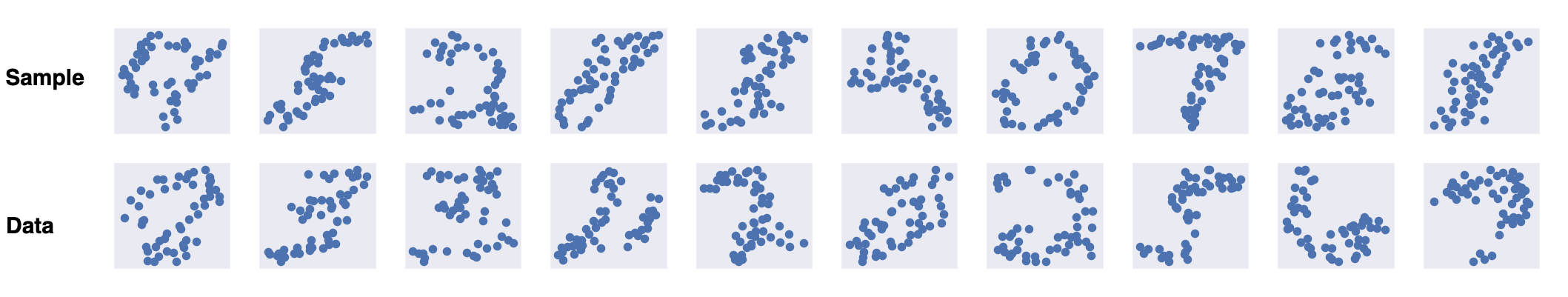}
    \includegraphics[width=0.95\linewidth]{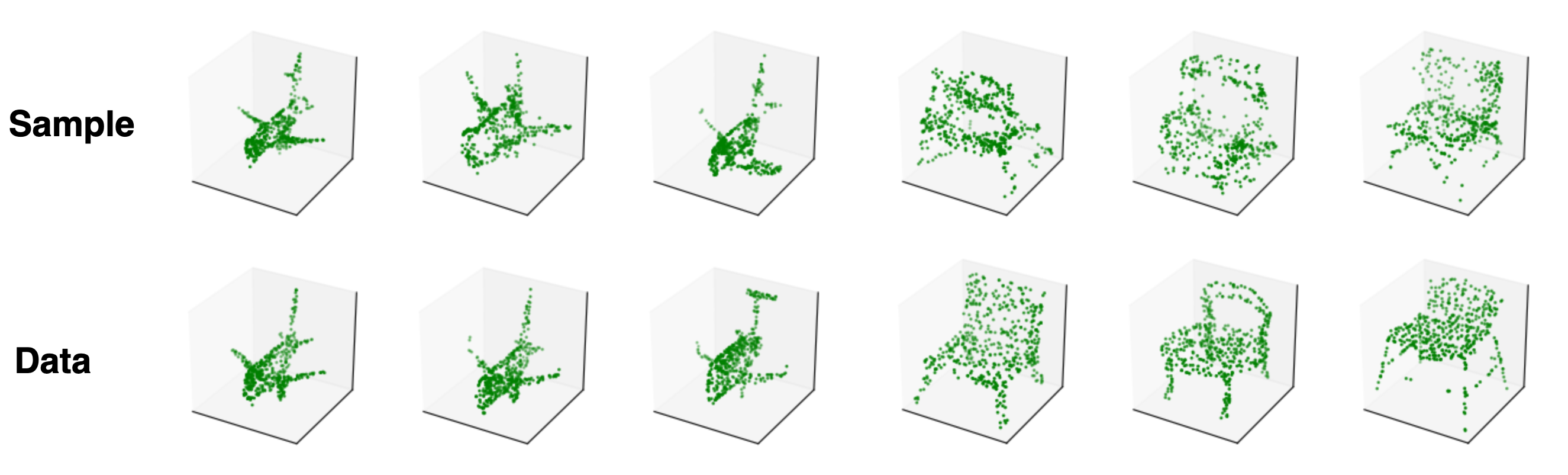}
    \caption{Generated samples and real data for SpatialMNIST (top) and ModelNet40 (bottom). SpatialMNIST consists of 50 points per set, and ModelNet40 contains 512 points.}
    \label{fig:set_gen}
\end{figure}

\vspace{-5pt}
\subsection{Set Generation and Density Estimation}
\label{subsec:set_gen}

\begin{figure}
    \centering
    \includegraphics[width=0.95\linewidth]{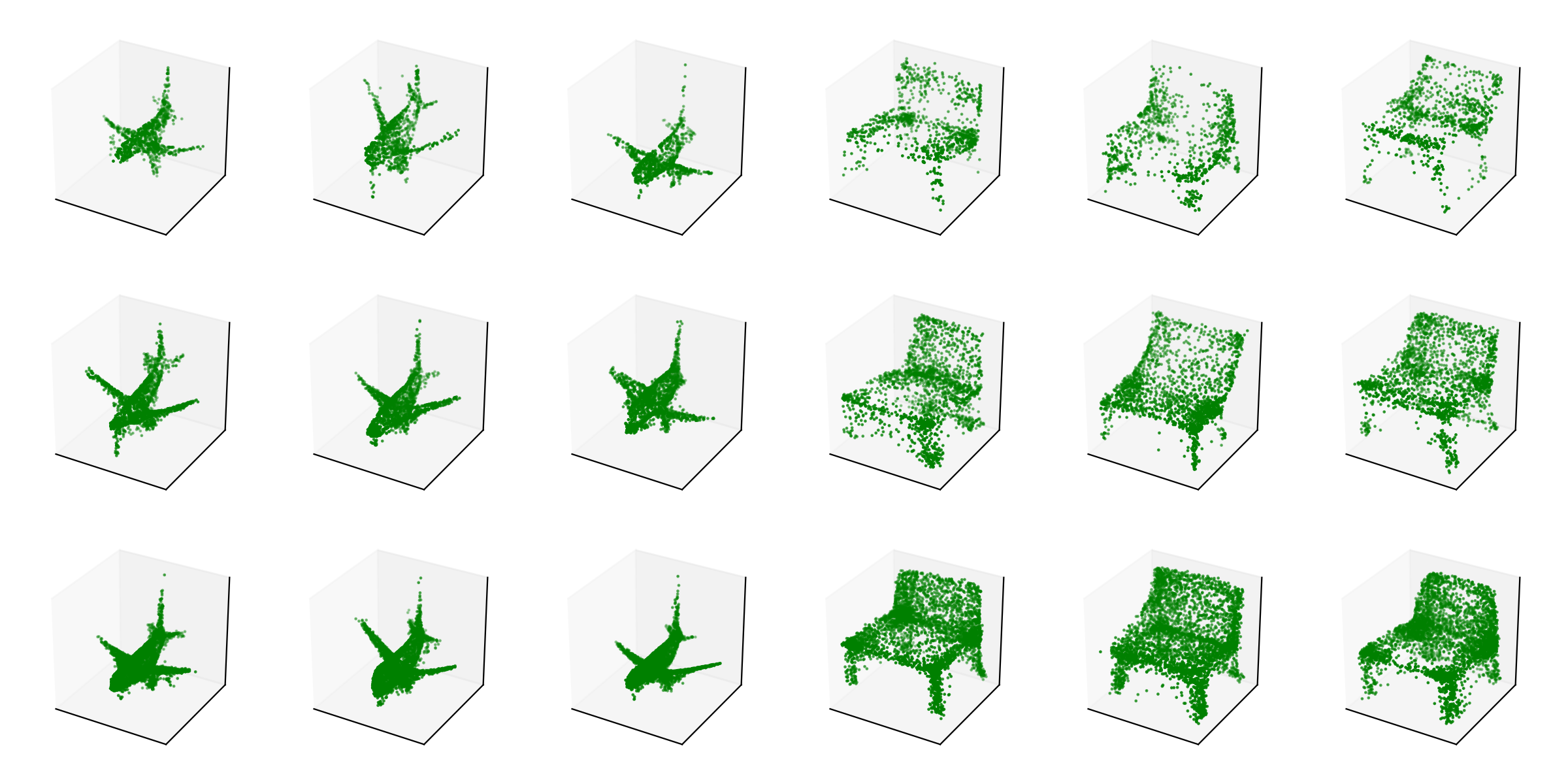}
    \caption{From top to bottom, we show sampled sets with 1024, 2048 and 4094 points, respectively. The model is trained with 512 point.}
    \label{fig:set_large}
\end{figure}

Next, we conduct experiments for set generative task using \textit{SpatialMNIST} \cite{edwards2016towards} and \textit{ModelNet40} \cite{edwards2016towards}. \textit{SpatialMNIST} consists of 50 2d points sampled uniformly from active pixels of MNIST.
\textit{ModelNet40} are constructed by sampling 512 points uniformly from one category.
Architectural details are provided in Appendix~\ref{sec:arch}.
The per-point log likelihood (PPLL) from the trained \method and other baselines can be found in Table \ref{tab:ppll}.
\method outperforms other models in all three datasets.
Figure \ref{fig:set_gen} shows the generated samples from our model.
Although our model is trained with only 512 points, it is capable of generating more points by sampling more points from the base distribution.
Figure \ref{fig:set_large} shows some examples for \textit{airplanes} and \textit{chairs}.

\begin{table}[ht]
    \caption{Per Point Log Likelihood (PPLL) on test set. Higher is better.}
    \label{tab:ppll}
    \centering
    \begin{tabular}{ccccc}
        \toprule Dataset & BRUNO \cite{bruno2018}  & NS \cite{edwards2016towards}  & FlowScan \cite{bender2020} & \method \\ \midrule
        Chairs  & 0.75 & 2.02 & 2.58 & \textbf{3.59} \\ 
        Airplanes & 2.71 & 4.09 & 4.81 & \textbf{5.13} \\ \midrule
        SpatialMNIST & -5.68 & -5.37 & -5.26 & \textbf{-5.21} \\ \bottomrule
    \end{tabular}
\end{table}

\vspace{-5pt}
\subsection{Temporal Set Modeling}\label{subsec:set_temp}

\begin{figure}[htb]
    \centering
    \includegraphics[width=\linewidth]{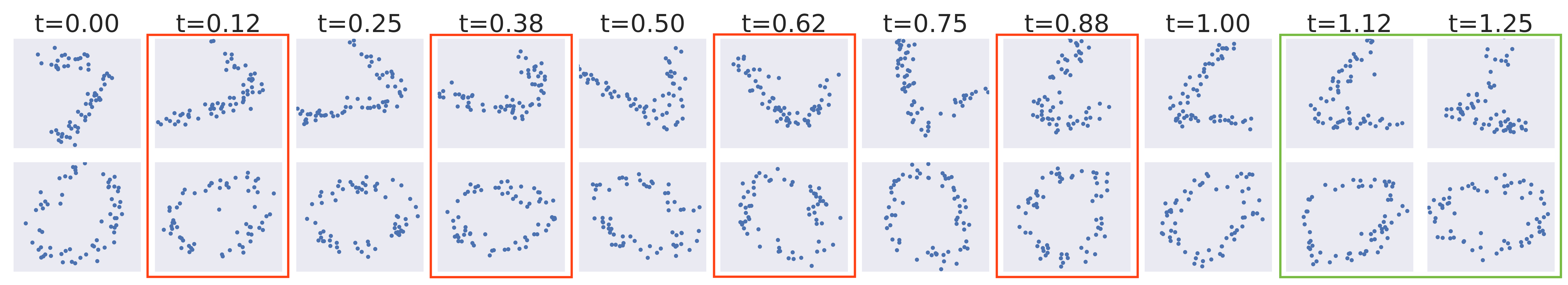} \\
    \caption{Samples form the temporal VAE. Red boxes indicate interpolated time steps, and green boxes indicate extrapolated time steps.}
    \label{fig:set_interp}
\end{figure}

For the set temporal generation task, we also use \textit{SpatialMNIST}.
To generate a temporal set series, we clockwise rotate the digits of MNIST dataset in a constant speed from $t=0$ to $t=1$ and sample 50 points from the active pixels at random.
We train our model at five \emph{fixed} time points, $t=[0,0.25,0.5,0.75,1]$.
For details about the architectures, please refer to appendix~\ref{sec:arch}.
Given the initial latent state, $z_0$, \method can generate latent state at any time and then generate corresponding samples conditioned on the latent state.
As shown in Fig.~\ref{fig:set_interp}, we can both interpolate and extrapolate to unseen time steps.
The samples generated at interpolated time maintain the smoothness over time.
See Appendix~\ref{sec:vae} for conditional samples where $z_0$ is encoded by a given series as a reconstruction.
Interpolation on latent code $z_0$ suggests our model learns a meaningful latent space. 

\section{Conclusion}
In this work, we extend neural ODEs to model exchangeable sets.
We prove that the solution of an ODE $\dot{x}(t) = f(x(t), t)$ is permutation equivariant w.r.t. initial value through time as long as its first order derivative $f(x(t), t)$ is permutation equivariant w.r.t. $x(t)$.
Therefore, we parameterize $f(x(t), t)$ as a permutation equivariant neural network and use black-box ODE solver to find the solution.
Since the ODE block is naturally invertible, we can apply our \method in a flow based generative model as an equivariant flow transformation.
According to the CNF formulation, we can compute the likelihood by solving an ODE. We also propose to model the time variant sets using a continuous-time VAE model. We observe smooth transition along time at both interpolated and extrapolated time steps.
In future works, we will evaluate on other applications, such as traffic tracking.

\section*{Broader Impact}
Making assessment over sets instead of instances gives us opportunity to leverage the dependencies over set elements. However, like any other models, it might unintentionally exploit the bias within the dataset. With this known issue, we encourage practitioners to carefully design the training set or utilize other debiasing techniques. In this work, we evaluate on point clouds of shape objects, which should not pose detrimental societal impact even if the learned dependencies does not reflect the actual ones.

Set generative models have the ability to generate fake data, which may incur ethical or legal issues when used improperly. There is urgent need to establish regulations and techniques to avoid misuse of the generated data.

\bibliographystyle{unsrt}
\bibliography{neurips_2020}

\clearpage

\appendix

\section{Proofs}\label{sec:proof}

\paragraph{Theorem 1}
(Permutation Equivariant ODE) Given an ODE $\dot{z}(t) = f(z(t), t), z(t)\in \mathcal{X}^n$ defined in an interval $[t_1, t_2]$. If function $f(z(t), t)$ is permutation equivariant w.r.t. $z(t)$, then the solution of the ODE, i.e., $z^\star(t), t\in[t_1, t_2]$ is permutation equivariant w.r.t. the initial value $z(t_1)$. We call the ODE with permutation equivariant properties as ExODE.

\begin{proof}
For any permutation $\pi(\cdot)$, we have
\begin{align*}
    \pi(z^\star (t)) &\overset{(1)}{=} \pi(z(t_1)) + \pi( \int_{t_1}^t f(z(\tau), \tau) d\tau ) \\
    &\overset{(2)}{=}  \pi(z(t_1)) + \int_{t_1}^t \pi(f(z(\tau), \tau)) d\tau \\
    &\overset{(3)}{=} \pi(z(t_1)) + \int_{t_1}^t f(\pi(z(\tau)), \tau) d\tau \\
    &\overset{(4)}{=} g(\pi(z(t_1)), f, t)
\end{align*}

\end{proof}

% \subsection{Extra Samples}

% \begin{figure}[ht]
%     \centering
%     \includegraphics[width=0.45\linewidth]{spmnist_appdx.png}\\
%     \includegraphics[width=0.45\linewidth]{airplane_appdx.png} ~~~
%     \includegraphics[width=0.45\linewidth]{chair_appdx.png}
%     \caption{Additional generated samples, SpatialMNIST on top, airplanes and chairs of ModelNet40 on bottom.}
%     \label{fig:spmnist_app}
% \end{figure}

\section{Temporal Set Modeling}\label{sec:vae}

\begin{figure}[ht!]
    \centering
    \includegraphics[width=0.8\textwidth]{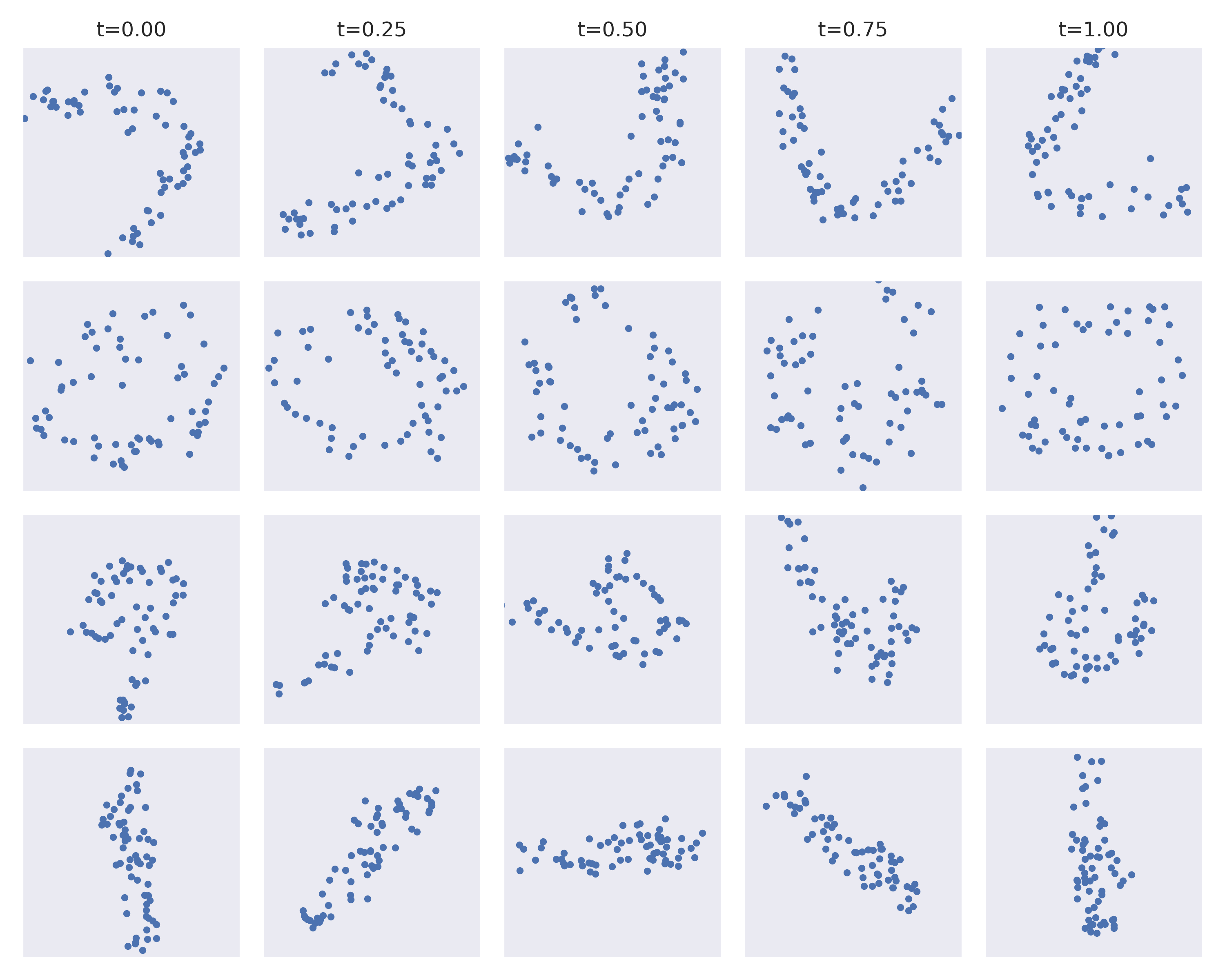}
    \caption{Additional samples from our temporal VAE.}
    \label{fig:uncond_sample}
\end{figure}

\begin{figure}[ht!]
    \centering
    \includegraphics[width=0.8\textwidth]{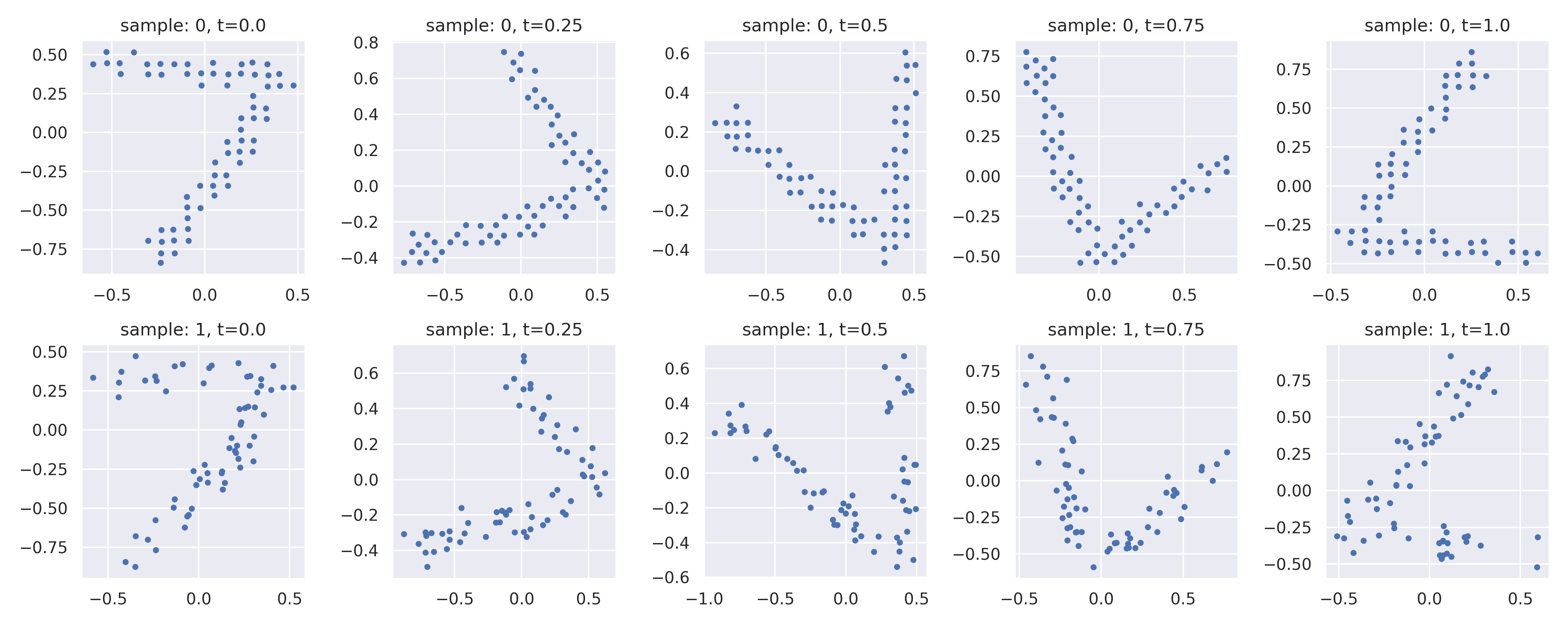}
    \caption{Conditional samples using the encoded $z_0$ of the first row.}
    \label{fig:cond_sample}
\end{figure}

\begin{figure}[ht!]
    \centering
    \includegraphics[width=0.8\textwidth]{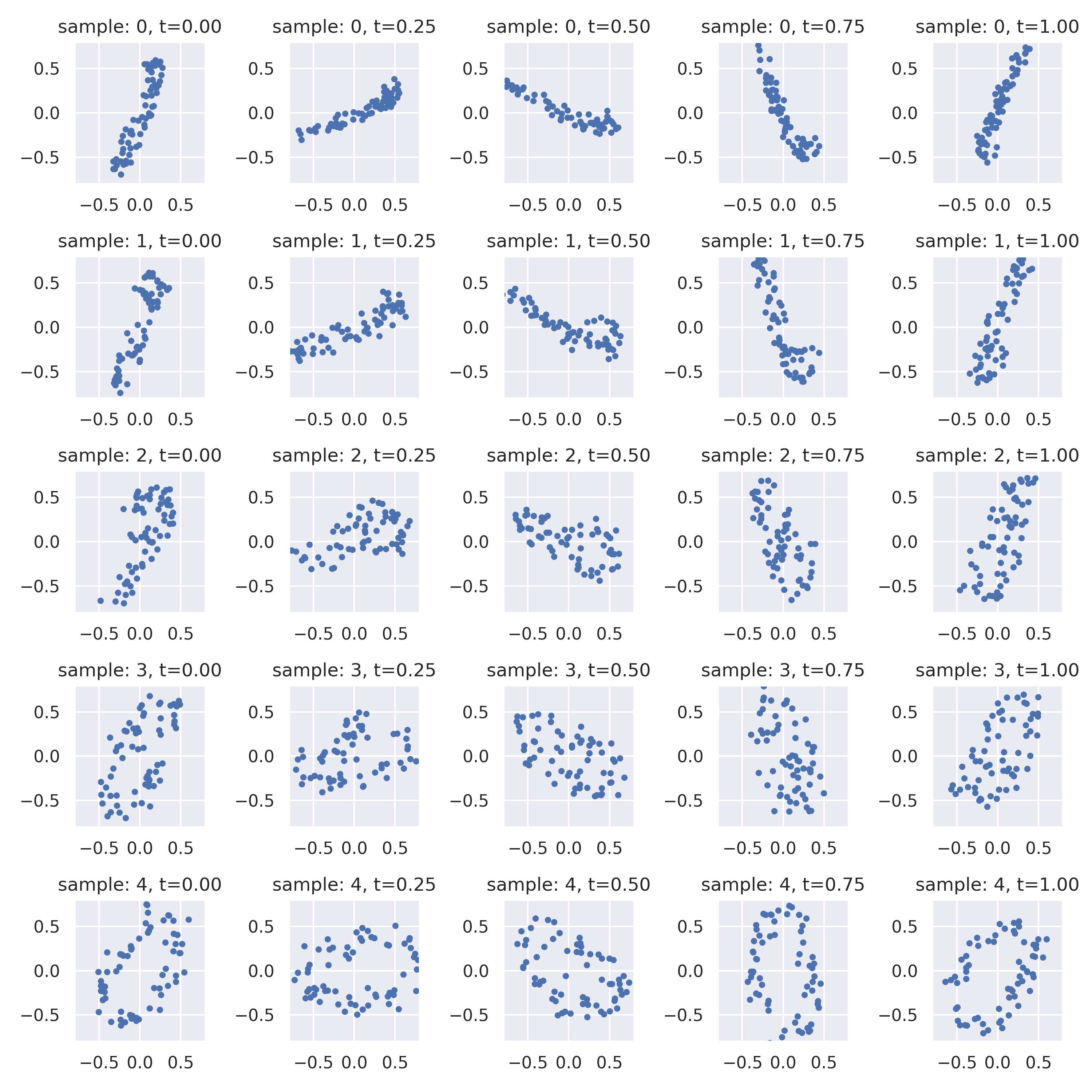}
    \caption{Interpolate $z_0$ from two different temporal sets (the first and the last row).}
    \label{fig:vae_interp}
\end{figure}

\section{Training details}

\subsection{Point cloud classification}

The details of network architecture we used are shown in Table \ref{tab:arch}. All the models are trained using Adam optimizer with $\beta_1=0.9$ and $\beta_2=0.999$. The learning rate and batch size are set to 1e-3 and 64 in all experiments. We use the fourth order Runge-Kutta solver to solve the \method in our model, and the numeric tolerance is set to 1e-5 in all experiments. We train our model on a single NVIDIA Tesla V100 GPU. For generalization, we randomly rotate and scale each set during training with $n=1000$ points.

\subsection{Set generation}
The details of network architecture are provided in Table \ref{tab:arch}. The batch is set to 128 in all experiments. We train our model using Adam optimizer with an initial learning rate of 1e-3 which we decay by a factor of 0.5 every 100 epochs. We use \texttt{dopri5} solver to solve the ODE with numeric tolerance of 1e-5.

\subsection{Set temporal model}
The dimension of the latent state variable $z_0$ is set to 128. We randomly sample 64 points uniformly from active pixels of MNIST dataset as a set. We train our model using Adam optimizer with learn rate 1e-3, $\beta_1=0.9$ and $\beta_2=0.999$, respectively. The batch size is set to 128 in all experiments. We use \texttt{dopri5} solver to solve the ODE used in our model, and the relative and absolute numeric tolerance are set to 1e-3 and 1e-4, respectively. All the models are trained on a single NVIDIA Tesla V100 GPU.

\paragraph{Decoder} The decoder models the reconstruction likelihood $p(\mathbf{x}_{t_i}|z_{t_i})$. We share the same decoder at different time. The \texttt{concatsquash}-like linear layers are used in our CNF decoder:

$$ CCS(\mathbf{x}, z, t) = (W_x\mathbf{x}+b_x) * \mathrm{gate} + \mathrm{bias}, $$

where $\mathrm{gate}=\sigma(W_{tt}t+W_{tz}z+b_t)$ and $\mathrm{bias}=W_{bt}t+W_{bz} z +b_bt$. In our experiment, we stack four \texttt{concatsquash} linear layers to model the dynamics $g_{\theta_d}$. We also use \texttt{Tanh} activation to connect the consecutive \texttt{concatsquash} linear layers. For more details of network architecture used in our model, see Table \ref{tab:arch}.

\section{Architecture}\label{sec:arch}
See next page.

\begin{table}[ht]
\caption{Detailed network architecture used in our experiments for different tasks.}
\label{tab:arch}
\centering
\begin{tabular}{llll}
\toprule \textbf{Model} & \textbf{Dataset} & \multicolumn{2}{c} { \textbf{Architecture} } \\
\midrule PointCloud Classification & ModelNet40 & Input & 64$\times$3$\times$100 or 1000 \\
(deepset block) & & FE & Conv1d 64x1(stride 1)  BN(64) Tanh \\
& &    & Conv1d 256x1(stride 1) BN(256) Tanh \\
& & \method & FC (512) Tanh FC(512) FC(256) \\
& & Pooling & Max(1) Flatten \\
& & Prediction & FC(128) BN(128) Tanh FC(40) \\
\midrule PointCloud Classification & ModelNet40 & Input & 64$\times$3$\times$100 or 1000 \\
(transformer block) & & FE & Conv1d 64x1(stride 1)  BN(64) Tanh \\
& &    & Conv1d 256x1(stride 1) BN(256) Tanh \\
& & \method & K: FC(256) Tanh FC(256) \\
& &       & Q: FC(256) Tanh FC(256) \\
& &       & V: FC(256) Tanh FC(256) \\
& &       & FC(256) \\
& & Pooling & Max(1) Flatten \\
& & Prediction & FC(128) BN(128) Tanh FC(40) \\
\midrule Set Generation & SpatialMNIST & Input & 128$\times$50$\times$2  \\
& & \method$\times$12 & K: FC(128) Tanh FC(128) Tanh FC(128) \\
& &       & Q: FC(128) Tanh FC(128) Tanh FC(128) \\
& &       & V: FC(128) Tanh FC(128) Tanh FC(128) \\
& &       & FC(2) \\ 
\midrule Set Generation & ModelNet40 & Input & 128$\times$512$\times$2  \\
& & \method$\times$12 & K: FC(128) Tanh FC(128) Tanh FC(128) \\
& &       & Q: FC(128) Tanh FC(128) Tanh FC(128) \\
& &       & V: FC(128) Tanh FC(128) Tanh FC(128) \\
& &       & FC(3) \\ 
\midrule Temporal Set Model & SpatialMNIST & Input & 128$\times$64$\times$2 \\
& & Encoder($\phi$) & Conv1d 128x1(stride 1) BN(128) ReLU \\
& & & Conv1d 128x1(stride 1) BN(128) ReLU \\
& & & Conv1d 256x1(stride 1) BN(256) ReLU \\
& & & Conv1d 512x1(stride 1) BN(512) Max \\
& & RNN & GRU(513, 512) (Concat $\Delta t$ as Input) \\
& & RNN\_to\_$z_0$ & mean: FC(256) BN(256) RELU \\
& & & FC(128) BN(128) RELU FC(128) \\
& & & std: FC(256) BN(256) ReLU \\
& & & FC(128) BN(128) ReLU FC(128) Exp \\
& & Latent: $z_0$ & 128 \\
& & ODE($z_t$) & FC(256) Tanh FC(256) Tanh FC(128)  \\
& & ODE($\hat{\mathbf{x}}_t$) & Concatsquash Linear $\times$ 4: \\
& & & 1) FC(2, 512) gate: FC(129, 512, bias=F) \\
& & & bias: FC(129, 512) (Concat $t$ and $z$) \\
& & & 2) FC(512, 512) gate: FC(129, 512, bias=F) \\
& & & bias: FC(129, 512) (Concat $t$ and $z$) \\
& & & 3) FC(512, 512) gate: FC(129, 512, bias=F) \\
& & & bias: FC(129, 512) (Concat $t$ and $z$) \\
& & & 4) FC(512, 2) gate: FC(129, 2, bias=F) \\
& & & bias: FC(129, 2) (Concat $t$ and $z$) \\
\bottomrule
\end{tabular}
\end{table}

\end{document}